\documentclass[%
 reprint,
 amsmath,amssymb,
pra,
]{revtex4-2}

\usepackage{graphicx}
\usepackage{dcolumn}
\usepackage{bm}

\begin{document}

\preprint{APS/123-QED}

\title{Prediction of Unobserved Bifurcation by Unsupervised Extraction of Slowly Time-Varying System Parameter Dynamics from Time Series Using Reservoir Computing}

\author{Keita Tokuda}
 \affiliation{Faculty of Health Data science, Juntendo University, 6-8-1, Hinode, Urayasu, Chiba, 279-0013, Japan.}
 \email{k.tokuda.jm@juntendo.ac.jp}


\author{Yuichi Katori}%
 \affiliation{
The School of Systems Information Science, Future University Hakodate, 116-2 Kamadanakano-cho, Hakodate, Hokkaido 041-8655, Japan}

\date{\today}

\begin{abstract}
Nonlinear and non-stationary processes are prevalent in various natural and physical phenomena, where system dynamics can change qualitatively due to bifurcation phenomena. Traditional machine learning methods have advanced our ability to learn and predict such systems from observed time series data. However, predicting the behavior of systems with temporal parameter variations without knowledge of true parameter values remains a significant challenge. This study leverages the reservoir computing framework to address this problem by unsupervised extraction of slowly varying system parameters from time series data. We propose a model architecture consisting of a slow reservoir with long timescale internal dynamics and a fast reservoir with short timescale dynamics. The slow reservoir extracts the temporal variation of system parameters, which are then used to predict unknown bifurcations in the fast dynamics. Through experiments using data generated from chaotic dynamical systems, we demonstrate the ability to predict bifurcations not present in the training data. Our approach shows potential for applications in fields such as neuroscience, material science, and weather prediction, where slow dynamics influencing qualitative changes are often unobservable.

\end{abstract}

\maketitle

\section{Introduction}

Nonlinear, non-stationary processes are abundant in various natural and physical phenomena. For instance, the dynamics of neurons are known to be strongly dependent on the state of the brain, determined by varying levels of attention, arousal, anesthesia, and sleep depth, as well as on different behavioral patterns like movement (\cite{Steriade3252,Buzski2002,Tokuda2019,Deco10.3389/fnsys.2020.00020}). Similarly, the response of physical systems can qualitatively change due to bifurcation phenomena as sample properties or experimental conditions vary (\cite{Bnard1901,Ertl1991,Itoh1996,Raab2023}). Various mathematical frameworks have been proposed to model non-stationary dynamics (\cite{WADDINGTON1961257,Kaneko2003,Rabinovich_PhysRevLett.87.068102,Katori2011, patel2021using}). One plausible and simple depiction is that system parameters vary over time or in different contexts (\cite{patel2021using})

Consider either a discrete nonlinear dynamical system:
\begin{eqnarray}
    \bm{x}(n+1) = f(\bm{x}(n);\lambda),
\end{eqnarray}
or a continuous dynamical system: 
\begin{eqnarray}
\label{dxdt}
\frac{\mathrm{d}\bm{x}}{\mathrm{d}t} = f(\bm{x};\lambda),
\end{eqnarray}
where $\bm{x} \in \mathbb{R}^n$ represents the dynamical variable expressing fast dynamics, and $\lambda$ is a parameter of function $f$ whose value can potentially lead to bifurcation in the dynamics of $\bm{x}$. In the context of modeling static nonlinear systems with a fixed value of $\lambda$, recent advancements in machine learning have enabled the rules governing the underlying system to be extracted and learned from observed time series data with much higher accuracy than before. In particular, by learning from time series data, reservoir computing has facilitated the creation of autonomous dynamical systems within the model that can generate time series resembling those of the target system, achieving high accuracy even in challenging problems such as learning chaotic systems. Furthermore, recent studies have demonstrated the prediction of unobserved bifurcations that are not present in the learning data (\cite{Kong_PhysRevResearch.3.013090, patel2021using,kim2021teaching,YoshitakaItoh2023}). In their settings, they have succeeded in predicting unknown bifurcations that occur when the parameter $\lambda$ takes values other than those used when generating the observed data. For example, Patel et al. addressed the bifurcation parameter $\lambda$ of a chaotic dynamical system not as static value but as a variable changing very slowly over time, and learned the time series generated by this system. After learning the one-step-ahead prediction task, they added a feedback loop to the reservoir, creating a closed-loop model that can generate time series as an autonomous dynamical system. They showed that, although learning the time series of $x$ using the conventional reservoir computing framework alone does not predict unobserved bifurcations, successful learning can be achieved by separately providing the reservoir with the true value of the parameter at each moment as an additional input. When the parameter values inputted during the prediction phase were different from those during learning, the model was able to predict bifurcations not included in the training data. Kim et al. demonstrated that the emergence of a Lorenz attractor not present in the training data could be predicted by first inputting time series generated from the Lorenz equations along with the true bifurcation parameter values into the reservoir, then forming a closed-loop model to create an autonomous dynamical system, and finally changing the input parameter values. These studies indicate that predicting unknown bifurcation phenomena is possible by additionally inputting the value of the bifurcation parameter into the reservoir. This suggests that the reservoir computing framework is capable of learning not just specific dynamical systems but families of dynamical systems, insinuating the potential to predict the emergence of system states qualitatively different from those observed in real data. However, these prior studies assume that the true value of the parameter is known, which is not the case in most real-world scenarios, including in brain data observation. Therefore, the question arises whether the behavior of non-stationary systems with temporal parameter variations can be predicted solely from observed time series data.

Various methods, including recurrence plots, supervised learning (\cite{PhysRevResearch.6.013196}), slow feature analysis(\cite{slowfeatureanalysis,ANTONELO2012178}), and hierarchical structures (\cite{Yonemura2021, katori,GALLICCHIO201787,Tamura2019}), have been reported for extracting the slowly moving components of system dynamics. In this study, we leverage the reservoir computing framework to address this problem. Our central idea is based on the following consideration: in a typical scenario, a reservoir receives a signal derived from a nonlinear dynamical system, such as one variable of the state vector $\bm{x}$ \textemdash \  e.g., $x_1$ \textemdash, in one step and predicts its value in the next time step. Previous studies have indicated that establishing generalized synchronization between the reservoir and the original system generating the input signal is crucial for achieving accurate predictions (\cite{PhysRevE.51.980,Caroll_10.1063/1.5128898,Ott_Attractor_reconstruction_10.1063/1.5039508,Lu_invertibleGeneralizedSync10.1063/5.0004344}), where generalized synchronization refers to the condition that the listening reservoir's state, $\mathbf{u}(t)$, is a continuous function, $\Psi(\bm{x})$, of the state of the state of the original system, $\bm{x}$. Especially, if the function $\Psi(\bm{x})$ is invertible, the reservoir's state $\mathbf{u}(t)$ has all the information about $\bm{x}$. It is reasonable to predict the value of another element \textemdash \   e.g., $x_2$ \textemdash, from partial observation of the system \textemdash \   e.g., only $x_1$ \textemdash, if generalized synchronization is established between the original system and the reservoir (\cite{Ott_Lu_unmeasured_10.1063/1.4979665}). Now, considering the parameter $\lambda$ varies slowly over time as expressed in equation \ref{dxdt}, the following system can be formulated:
\begin{eqnarray} 
\left\{ 
  \begin{alignedat}{2}   
    \frac{\mathrm{d}\bm{x}}{\mathrm{d}t} &= f_x(\bm{x}; \lambda) 
    \\ 
    \frac{\mathrm{d}\lambda}{\mathrm{d}t} &= f_{\lambda}(\bm{x}, \lambda)
  \end{alignedat} 
\right .
\end{eqnarray}
Let $X$ be a concatenation of $\bm{x}$ and $\lambda$, defined as $X = \, {}^\mathrm{t}\! \left( \, {}^\mathrm{t}\!\bm{x}, \lambda \right)$, then this system can be represented as a single ordinary differential equation (ODE):
\begin{equation}
\frac{\mathrm{d}\bm{X}}{\mathrm{d}t} = F(\bm{X}).
\end{equation}
We posit that the signal is generated from the trajectory of this concatenated system’s attractor. When the signal originating from $\bm{x}$ is input into the reservoir and invertible generalized synchronization between the reservoir state $\bm{u}$ and $\bm{X}$ is achieved, the reservoir's state has full information about $\lambda$. While above discussion is speculative, previous studies have shown that by adjusting the reservoir's timescale and structure, the reservoir can successfully extract the slow dynamics of the signal source system \cite{Manneschi,Jaeger2008DiscoveringMD,GALLICCHIO201787,Gouhei,Yonemura2021}). The extraction of such slow or static system states within the reservoir computing framework, where internal couplings are not altered during learning, suggests that unsupervised extraction of such information is possible using reservoirs. We first aim to verify whether it is possible to extract the true variation of parameter $\lambda$'s by simply extracting the slowly varying variables within the reservoir (section \ref{esf} Experiment 1).

Patel et al. have demonstrated that predicting the time series of the concatenated system $X$ cannot be achieved by a simple single reservoir (\cite{patel2021using}). The challenges addressed in this paper are twofold: (1) estimating the unobservable slowly varying parameter values (section \ref{esf} Experiment 1), and (2) predicting unknown bifurcations in the fast dynamics under the variation of such parameters (section \ref{pub} Experiment 2). While the second challenge has been tackled by Patel et al. and Kim et al. in scenarios where the true parameter value is known, in this study, we explore the possibility of learning from observational data generated by nonlinear systems and predicting unknown bifurcations without the knowledge of true parameter values. We allow the bifurcation parameter values to change over time but assume these changes occur on a significantly longer timescale compared to the system's fast dynamics. Previous studies suggest that extracting the slowly changing parameter values from time series observations in an unsupervised manner may allow us to substitute the true parameter value with an estimated one.

The architecture of the model proposed in this study comprises two types of reservoirs stacked in layers: a slow reservoir with long timescale internal dynamics and a fast reservoir with short timescale dynamics. Assuming a nonlinear system with a very slowly changing bifurcation parameter value as the signal source, we input observational data obtained from the fast dynamics into these reservoirs. We found that when the variables that change slowly are extracted from the internal state of the slow reservoir, they trace the temporal variation of the system's parameter. Although the variables extracted from the slow reservoir differ in amplitude scale from the true parameter values, we show that adding these variables and the observational time series to the fast reservoir allows for the prediction of unknown bifurcations, resembling the true parameter values provided in prior studies.

\section{Materials and Methods}

\subsection{Problem Setting}
Consider the following nonlinear differential equations:
\begin{eqnarray} 
\label{problem}
\left\{ 
  \begin{alignedat}{2}   
    \frac{\mathrm{d}\bm{x}}{\mathrm{d}t} &= f_x(\bm{x}; \lambda) 
    \\ 
    \frac{\mathrm{d}\lambda}{\mathrm{d}t} &= f_{\lambda}(\lambda, t)
  \end{alignedat} 
\right.
\end{eqnarray}
As implied by this equation, the parameter $\lambda$ is assumed to vary over time. However, in this paper, we sometimes do not explicitly define $f_{\lambda}$ and, instead, assume $\lambda$ is simply a function of $t$. In either case, the temporal change of $\lambda$ is assumed to be significantly  slower than that of $\bm{x}$. Herein, we consider concrete examples of $f_x$ by examining numerical computations derived from the Lorenz and R\"ossler equations.

We assume the observation time series $y$ is given as a function of the fast variable $\bm{x}$, as follows:
\begin{eqnarray}
    y(n) = g(\bm{x}(\Delta t \cdot n)))
\end{eqnarray}
where $g(\bm{x}(t))$ is a function of $\bm{x}(t)$. We assume $g$ to be well-behaved, such as a smooth and differentiable function, but without including the full observation of the state $\bm{x}$, i.e., $\mathrm{dim}\ y < \mathrm{dim}\ \bm{x}$. In this paper, we use 
\begin{eqnarray}
    g(\bm{x}) = x_1,
\end{eqnarray}
where $x_1$ is the first element of the vector $\bm{x}$. We naturally suppose that the time series are obtained by temporally discretizing the continuous signal with a specific time step, $\Delta t$. The problem we address here is whether detecting changes in the slowly varying parameter $\lambda$ and predicting unknown bifurcations is possible based on the observed $y(n)$.

\begin{figure}[h]
\centering
    \begin{minipage}[b]{85 mm}
    \includegraphics[width=85 mm]{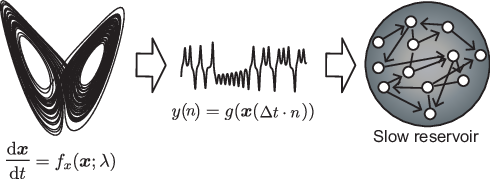}
    \caption{Schematic diagram of the numerical examination discussed in Section \ref{esf}. The observation signal $y(n)$ is generated by a nonlinear system $\mathrm{d}x/\mathrm{d}t = f_x (\bm{x}; \lambda)$. The slow reservoir consists of leaky neuron models with a very long leak rate, and the spectral radius of the recurrent connection is set to 1.}
    \label{schematic1}
    \end{minipage}
\end{figure}

\subsection{Model Architecture}
\begin{figure*}
    \centering
    \includegraphics[width=134 mm]{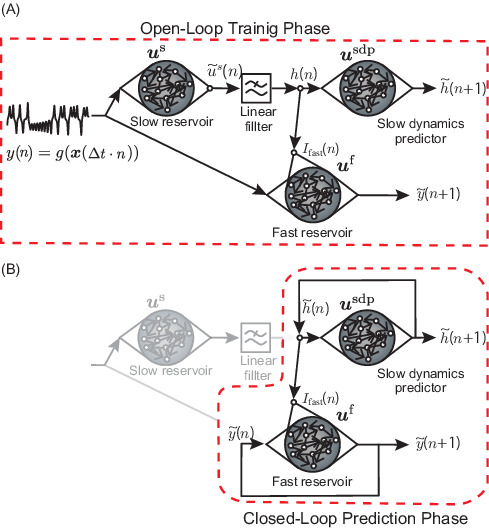}
    \caption{Schematic diagram showing the open-loop training phase and the closed-loop prediction phase. (A) In the training phase, observation $y(n)$ is fed to both the slow and fast reservoirs. The output of the slow reservoir is also input to the fast reservoir. Additionally, the output of the slow reservoir is input to the ``slow dynamics predictor'' reservoir. After the training phase, the output weights of both the slow dynamics predictor and the fast reservoir are optimized to conduct one-step-ahead prediction of their own inputs. (B) In the prediction phase, feedback loops are added to the slow dynamics predictor and the fast reservoir to make the whole system a single autonomous dynamical system that can predict time series of $y(n)$.}
    \label{schematic_unobserved}
\end{figure*}

In this study, we conduct two experiments: (1) estimating the unobservable slowly varying parameter values (section \ref{esf} Experiment 1), and (2) predicting unknown bifurcations in the fast dynamics under the variation of such parameters (section \ref{pub} Experiment 2).
In Experiment 1, as shown in fig.~\ref{schematic1}, we input time series observations generated from the attractor trajectory of a nonlinear dynamical system into a reservoir with a slow time constant and observe the internal state of the reservoir. We check whether there are nodes in the internal state that exhibit fluctuations similar to the slow movement of the parameter of the dynamical system generating the data. The experimental setup is shown in fig.~\ref{schematic1}. We refer to this reservoir with a slow time constant as the \textit{slow reservoir}. 
In Experiment 2, we test whether the movements of the parameters extracted from the slow reservoir and the observed time series can be used as inputs to predict bifurcations in the attractor. We refer to this downstream reservoir as the \textit{fast reservoir}. The model architecture is shown in fig.~\ref{schematic_unobserved}. In Experiment 2, during the training phase, the model learns from the time series, and afterwards, by introducing feedback, it operates as a fully autonomous system. The model must predict both the changes in the slow-moving parameter and the values of the fast dynamical variables. Therefore, the model includes a third reservoir, called the \textit{slow dynamics predictor}, which predicts the time series output of the slow reservoir (fig.~\ref{schematic_unobserved}). In the test phase after training, both the slow dynamics predictor and the fast reservoir are provided with feedback from their outputs, forming a \textit{closed-loop model}.

\subsection{Reservoir model}
\subsubsection{Slow reservoir}
We employed a reservoir consisting of leaky integrator neurons with long time constants to extract the slow variables of the system (\cite{JAEGER2007335}). Namely, the leak rate, $\alpha$, is set close to 1. To ensure that the reservoir dynamics arising from neuron interactions also exhibit long time constants, we adjust the spectral radius of the recurrent connection strength, $\bm{W}^{\mathrm{s}} \in \mathbb{R}^{N^{\mathrm{s}} \times N^{\mathrm{s}}}$ to be very close or equal to one.
The dynamics of the slow reservoir are defined by the following equation:
\begin{widetext}
\begin{eqnarray}
\label{slowreservoirequ}
\bm{u}^{\mathrm{s}}(n+1) &=& \alpha \bm{u}^{\mathrm{s}}(n) + (1 - \alpha) \tanh(\bm{W}^{\mathrm{s}} \bm{u}^{\mathrm{s}}(n) + \bm{W}^{\mathrm{s}}_{\text{in}} \cdot y(n) + \bm{b}^\mathrm{s}) 
\end{eqnarray}
\end{widetext}
where $\bm{u}^{\mathrm{s}} = (u_1^{\mathrm{s}}, u_2^{\mathrm{s}}, \ldots, u^{\mathrm{s}}_{N^{\mathrm{s}}})^T$ is the internal state of the reservoir, $\bm{b}^{\mathrm{s}} \in \mathbb{R}^{N^{\mathrm{s}}}$  is the bias term, $\bm{W}^{\mathrm{s}} \in \mathbb{R}^{N^{\mathrm{s}} \times N^{\mathrm{s}}}$ is the recurrent connection strength, $\bm{W}^{\mathrm{s}}_{\mathrm{in}} \in \mathbb{R}^{N^{\mathrm{s}} \times 1}$ is the input connection strength, $\alpha$ is the leak rate, $\tanh(x) = \frac{e^{2x} - 1}{e^{2x} + 1}$ is the  hyperbolic tangent, $N^{\mathrm{s}}$ is the number of neurons in the reservoir, and $y(n) \in \mathbb{R}$ is a one-dimensional time series observation derived from the data generation models mentioned earlier. The elements of $\bm{W}^{\mathrm{s}}$ are drawn from an  i.i.d. Gaussian normal distribution, and then $\bm{W}^{\mathrm{s}}$ is normalized by multiplying a constant factor so that the spectral radius $\rho (\bm{W}^{\mathrm{s}} )$ satisfies $\rho (\bm{W}^{\mathrm{s}} ) = 1$. The elements in the input connection matrix $\bm{W}^{\mathrm{s}}_{\mathrm{in}}$ and in vector $\bm{b}^{\mathrm{s}}$ are drawn from an  i.i.d. uniform distribution over intervals $[-\chi^{\mathrm{s}}_{\mathrm{in}}, \ \chi^{\mathrm{s}}_{\mathrm{in}}]$ and $[-\chi^{\mathrm{s}}_{\mathrm{b}}, \ \chi^{\mathrm{s}}_{\mathrm{ib}}]$, respectively. The parameter values used are $N^{\mathrm{s}} = 500$ and $\alpha = 0.995$. When applying the Lorenz system as the data generation model, we used $\chi^{\mathrm{s}}_{\mathrm{in}} = 0.5$ and $\chi^{\mathrm{s}}_{\mathrm{b}} = 5$, whereas when applying the R\"ossler model, we set $\chi^{\mathrm{s}}_{\mathrm{in}} = 15$ and $\chi^{\mathrm{s}}_{\mathrm{b}} = 150$. To determine the slow dynamics of the target system, we extracted slowly changing elements of the internal state $\bm{u}^{\mathrm{s}}$, which were heuristically selected using the following procedure:
\begin{itemize}
    \item For each $i$, calculate a moving average of the time series $u^{\mathrm{s}}_i(n)$ using a time window $n_{\mathrm{window}}$ with a specific width, where $u^{\mathrm{s}}_i(n)$ is the $i$th element of the internal state $\bm{u}^{\mathrm{s}}$. Let $\overline{u^{\mathrm{s}}_i}(n)$ denote this moving average:
    \begin{eqnarray}
        \overline{u^{\mathrm{s}}_i}(n) = \dfrac{\sum^{n_{\mathrm{window}}-1}_{k=0}  u^{\mathrm{s}}_i(n-k) }{n_{\mathrm{window}}}
    \end{eqnarray}
    \item For each $i$, calculate the fluctuation around its own moving average during the training phase using the standard deviation $SD\left[  u^{\mathrm{s}}_i(n) -\overline{u^{\mathrm{s}}_i}(n) \right]$.
    \item Choose the elements with the lowest fluctuation. In this study, the top 10\%, namely 50 nodes, were selected. Let $S_{\mathrm{slow}}^{10 \%} = \{i \mid u^{\mathrm{s}}_i \textrm{ is in the slowest } 10\% \textrm{ of nodes}\}$ denote this set.
    \item Calculate the instantaneous average of the absolute values of the selected elements:
    \begin{eqnarray}
        \tilde{u}^s(n) = \dfrac{1}{|S_{\mathrm{slow}}^{10 \%}|} \sum_{i \in S_{\mathrm{slow}}^{10 \%}}
        |u^{\mathrm{s}}_i(n) |,
    \end{eqnarray}
    where $\tilde{u}^s(n)$ denotes the extracted slow feature, and $|S_{\mathrm{slow}}^{10 \%}|$ is the cardinality of the set $S_{\mathrm{slow}}^{10 \%}$. The absolute value of each node is taken because some nodes exhibit changes that follow the same pattern as the parameter changes, while others show changes that are the inverse. Simply averaging these values would cancel them out, resulting in a near-zero $\tilde{u}^s(n)$. By taking the absolute value before averaging, we ensure that the contributions of all selected nodes are positively accounted for, avoiding this cancellation effect due to the central symmetry of the $\tanh$ function and the distribution of each weight element.
\end{itemize}

In the numerical experiment with the closed-loop model discussed in Section \ref{pub}, the extracted slow feature is further smoothed by a linear filter before being fed to the downstream reservoirs (namely, fast reservoir reservoir and the slow dynamics predictor) to stabilize the learning process. The filtering is described by the following linear dynamics:
\begin{eqnarray}
\label{linFil}
    h(n+1) = (1- \dfrac{1}{\tau_{\mathrm{f}}}) h(n) + \dfrac{\tilde{u}^s(n)}{\tau_{\mathrm{f}}},
\end{eqnarray}
where $\tau_{\mathrm{f}}$ is the time constant of the filter. This equation is derived from the following linear dynamics:
\begin{eqnarray}
    \tau_{\mathrm{f}}\dfrac{ h(n+\Delta n) - h(n) }{\Delta n}= -h(n) + \tilde{u}^s(n)
\end{eqnarray}
Equation \ref{linFil} is obtained by substituting $\Delta n = 1$. The parameter value used is $\tau_{\mathrm{f}}=200$. The application of a linear filter does not significantly alter the shape of the time series; it is used solely for removing high-frequency components and smoothing (fig.~S1).

\subsubsection{Fast reservoir}
In addition to the slow reservoir described above, which is employed to extract the slow components of the dynamics, we utilize another reservoir to capture the evolution laws of the fast dynamics of the target system (fig. \ref{schematic_unobserved}). The model is almost same as equation \ref{slowreservoirequ} but with two inputs and different parameter values:
\begin{widetext}
\begin{eqnarray}
\label{FastRes}
\bm{u}^{\mathrm{f}}(n+1) &=&  
\alpha \bm{u}^{\mathrm{f}}(n) +
(1-\alpha)
\tanh(\bm{W}^{\mathrm{f}} \bm{u}^{\mathrm{f}}(n) + \bm{W}^{\mathrm{f}}_{\text{in}} \cdot y(n) 
+ \bm{W}_{\mathrm{param}} \cdot   I_{\mathrm{fast}}(n)
+ \bm{b}^{\mathrm{f}}) 
\end{eqnarray}
\end{widetext}
where $\bm{u}^{\mathrm{f}} = (u_1, u_2, \ldots, u_{N^{\mathrm{f}}})^T$ is the internal state of the reservoir, $\bm{b}^{\mathrm{f}} \in \mathbb{R}^{N^{\mathrm{f}}}$ is the bias term, $\bm{W}^{\mathrm{f}} \in \mathbb{R}^{N^{\mathrm{f}} \times N^{\mathrm{f}}}$ is the recurrent connection strength, $\bm{W}^{\mathrm{f}}_{\mathrm{in}} \in \mathbb{R}^{N^{\mathrm{f}} \times 1}$ is the input connection strength, $\tanh(x) = \frac{e^{2x} - 1}{e^{2x} + 1}$ is the hyperbolic tangent, and $N^{\mathrm{f}}$ is the number of neurons in the reservoir. As in previous studies (\cite{patel2021using, kim2021teaching}), a slowly changing parameter value that acts as the bifurcation parameter is also fed to the reservoir, as expressed by the term $\bm{W}_{\mathrm{param}} \cdot I_{\mathrm{fast}}(n)$ in the RHS of the equation, where $\bm{W}_{\mathrm{param}} \in \mathbb{R}^{N^{\mathrm{f}} \times 1}$ is the input connection strength and $I_{\mathrm{fast}}(n)$ is the additional input to the fast reservoir receiving the slow component. Unlike previous studies, $I_{\mathrm{fast}}(n)$ is not the true parameter value of the target system but the output of either one of other reservoirs, the slow reservoir or the slow dynamics predictor described below (fig. \ref{schematic_unobserved}). A sparse matrix is used for $\bm{W}^{\mathrm{f}}$, such that randomly chosen 2\% of the edges are assigned non-zero values, whereas the rest are set to zero. The weight values of the 2\% edges are drawn from an i.i.d. uniform distribution over the interval $[0 \ \ 1]$, and $\bm{W}^{\mathrm{f}}$ is normalized so that the spectral radius $\rho (\bm{W}^{\mathrm{f}} )$ satisfies $\rho (\bm{W}^{\mathrm{f}}) = 0.95$. The elements in $\bm{W}^{\mathrm{f}}_{\mathrm{in}}, \bm{W}_{\mathrm{param}}$, and $\bm{b}^{\mathrm{f}}$ are drawn from an i.i.d. uniform distribution over intervals $[-\chi^{\mathrm{f}}_{\mathrm{in}}, \ \chi^{\mathrm{f}}_{\mathrm{in}}]$ , $[-\chi^{\mathrm{f}}_{\mathrm{param}}, \ \chi^{\mathrm{f}}_{\mathrm{param}}]$, and $[-\chi^{\mathrm{f}}_{\mathrm{b}}, \ \chi^{\mathrm{f}}_{\mathrm{b}}]$, respectively. The parameter values used are $N^{\mathrm{f}} = 2000$, $ \chi^{\mathrm{f}}_{\mathrm{in}}= 0.75$, $ \chi^{\mathrm{f}}_{\mathrm{param}} = 0.15$, $\chi^{\mathrm{f}}_{\mathrm{b}}=15$, and $\alpha = 0.95$.

\subsubsection{Slow dynamics predictor}
In Section \ref{pub}, we demonstrate the construction of a closed-loop model capable of predicting unobserved bifurcations without requiring an observation signal as its input (fig. \ref{schematic_unobserved}). Typically, in reservoir computing, a closed-loop model can be established by simply adding a feedback loop, using the reservoir's output as its input at the next time step. This approach is applied to the fast reservoir (fig. \ref{schematic_unobserved}). However, due to the extraction of slow dynamics from the observation of fast dynamics using the slow reservoir, the output of the slow reservoir exhibits different temporal properties and cannot be used as feedback to substitute the input at the next time step. Therefore, to construct a closed-loop model, we introduce the slow dynamics predictor, which is an additional reservoir that predicts the evolution of the slow component of the target dynamics. In the training phase, the slow component $I_{\mathrm{fast}}(n)$ in the RHS of equation \ref{FastRes} originates from the output of the slow reservoir, namely, $I_{\mathrm{fast}}(n) = h(n)$. In the prediction phase, the slow dynamics predictor works as an autonomous system by adding the closed loop, and we substitute $I_{\mathrm{fast}}(n)$ in the RHS of equation \ref{FastRes} with the prediction of this slow dynamics predictor, namely, $I_{\mathrm{fast}}(n) = \tilde{h}(n)$.  We use a standard echo state network for this reservoir (\cite{Jaeger2004}). The model is almost the same as equation \ref{FastRes} but with slightly different parameter values and without the input $y(n)$ and leak term. The dynamics of the slow dynamics predictor during the training phase is described as follows:
\begin{widetext}
\begin{eqnarray}
\bm{u}^{\mathrm{sdp}}(n+1) &=&  \tanh(\bm{W}^{\mathrm{sdp}} \bm{u}^{\mathrm{sdp}}(n) 
+ \bm{W}_{\mathrm{param}} \cdot  h(n)
+ \bm{b}^{\mathrm{sdp}}) .
\end{eqnarray}
\end{widetext}
The parameter values used are $N^{\mathrm{sdp}} = 500$ and $ \  \chi^{\mathrm{sdp}}_{\mathrm{param}}=\chi^{\mathrm{sdp}}_{\mathrm{b}}= 5 \times 10^{-3}$.

\subsection{Training}
Figure \ref{schematic_unobserved} describes the model architecture used for the one-step-ahead prediction task. Training is conducted using "teacher forcing," where, during the training phase, the observation time series to be predicted serves as the external force that drives the reservoir (fig. \ref{schematic_unobserved} (A)). To train the fast reservoir, the sum of squared output errors in one-ahead-prediction plus the regularization term is minimized with respect to $\bm{W}_{\mathrm{out}}^{\mathrm{f}}$:
\begin{eqnarray}
    \sum_n \left( 
    y(n+1) -
    \bm{W}_{\mathrm{out}}^{\mathrm{f}}  \bm{u}^{\mathrm{f}}(n+1)
    \right)^2 
    +
    \beta |\bm{W}_{\mathrm{out}}^{\mathrm{f}}|_{\mathrm{fro}}
\end{eqnarray}
where $|\bm{W}_{\mathrm{out}}^{\mathrm{f}}|_{\mathrm{fro}}$ is the Frobenius norm of matrix $\bm{W}_{\mathrm{out}}^{\mathrm{f}}$, $\beta$ is the regularization coefficient, $\bm{W}_{\mathrm{out}}^{\mathrm{f}}$ is the output connection strength. Here, the Frobenius norm \( \|A\|_F \) of a matrix $A=(a_{ij})$ refers to the square root of the sum of the absolute squares of its elements as follows:
\begin{eqnarray}
\|A\|_F = \sqrt{\sum_{i=1}^{m} \sum_{j=1}^{n} |a_{ij}|^2}
\end{eqnarray}
Similarly, to train the slow dynamics predictor , the sum of squared output errors in one-ahead-prediction plus the regularization term is minimized with respect to $\bm{W}_{\mathrm{out}}^{\mathrm{sdp}}$:
\begin{eqnarray}
\nonumber
    \sum_n \left( 
    h(n+1) -
    \bm{W}_{\mathrm{out}}^{\mathrm{sdp}}  \bm{u}^{\mathrm{sdp}}(n+1)
    \right)^2 
    \\ 
    +
    \beta |\bm{W}_{\mathrm{out}}^{\mathrm{sdp}}|_{\mathrm{fro}}.
\end{eqnarray}

The training of the fast reservoir and the slow dynamics predictor take place parallel after the training phase.

\subsection{Closed-loop model}
Following the training phase, feedback loops are incorporated into the slow dynamics predictor and the fast reservoir, transforming the entire system into a single autonomous dynamical system capable of generating predictions for the time series $y(n)$ (fig. \ref{schematic_unobserved} (B)). The fast reservoir with the feedback loop is described by the following equations:
\begin{widetext}
\begin{eqnarray}
\label{FastRes2}
\left\{ 
\begin{aligned}
\bm{u}^{\mathrm{f}}(n+1) &=  \tanh(\bm{W}^{\mathrm{f}} \bm{u}^{\mathrm{f}}(n) + \bm{W}^{\mathrm{f}}_{\text{in}} \cdot \tilde{y^{\mathrm{f}}}(n) 
+ \bm{W}_{\mathrm{param}} \cdot  \tilde{h}(n)
+ \bm{b}^{\mathrm{f}}) 
 \\
\tilde{y^{\mathrm{f}}}(n+1)  &= \bm{W}_{\mathrm{out}}^{\mathrm{f}} \bm{u}^{\mathrm{f}}(n+1)
\end{aligned} 
\right.
\end{eqnarray}
\end{widetext}
where $\tilde{h}(n)$ is the external input to the system, whose evolution is governed by the slow dynamics predictor model described as follows:
\begin{widetext}
\begin{eqnarray}
\label{srp}
\left\{ 
\begin{aligned}
\bm{u}^{\mathrm{sdp}}(n+1) &=  \tanh(\bm{W}^{\mathrm{sdp}} \bm{u}^{\mathrm{sdp}}(n) 
+ \bm{W}_{\mathrm{param}} \cdot  \tilde{h}(n)
+ \bm{b}^{\mathrm{sdp}}) 
 \\
\tilde{h}(n+1)  &= \bm{W}_{\mathrm{out}}^{\mathrm{sdp}} \bm{u}^{\mathrm{sdp}}(n+1)
\end{aligned} 
\right.
\end{eqnarray}
\end{widetext}
Equations \ref{FastRes2} and \ref{srp} collectively form the autonomous dynamical system capable of independently generating a time series.

\subsection{Largest Lyapunov exponent estimation}

After the training phase, the largest Lyapunov exponent (LLE) is computed for the fast reservoir with feedback (eq.~\ref{FastRes2}). In Experiment 2, both during the training and prediction phases, the fast reservoir receives the time-varying output, namely the smoothed output of the slow reservoir, $h(n)$, or the output of the slow dynamics predictor, $\tilde{h}(n)$, as its input $I_{\mathrm{fast}}(n)$. Here we calculate the LLE of the fast reservoir by fixing the value of $I_{\mathrm{fast}}(n)$. Namely, the LLE of the fast reservoir with the parameter $n$ is defined by the LLE of the following dynamical system:
\begin{widetext}
\begin{eqnarray}
	\label{autonomousFastReservoir0}
\left\{
\begin{aligned}
\bm{u}^{\mathrm{f}}(k+1) &=  \tanh(\bm{W}^{\mathrm{f}} \bm{u}^{\mathrm{f}}(k) + \bm{W}^{\mathrm{f}}_{\text{in}} \cdot \tilde{y^{\mathrm{f}}}(k) 
+ \bm{W}_{\mathrm{param}} \cdot  I_{\mathrm{fast}}(n)
+ \bm{b}^{\mathrm{f}}) 
\\	
\tilde{y^{\mathrm{f}}}(k+1) &= \bm{W}_{\mathrm{out}}^{\mathrm{f}} \bm{u}^{\mathrm{f}}(k+1)
\end{aligned}
\right.
\end{eqnarray}
\end{widetext}
where $k$ denotes the time step and $n$ is regarded as a constant value. Substituting the second expression of eq.~\ref{autonomousFastReservoir0} into the first one yields the autonomous dynamical system with the parameter $I_{\mathrm{fast}}(n)$ as follows:
\begin{widetext}
\begin{eqnarray}
\label{noy}
\bm{u}^{\mathrm{f}}(k+1) &=  \tanh(\bm{W}^{\mathrm{f}} \bm{u}^{\mathrm{f}}(k) + \bm{W}^{\mathrm{f}}_{\text{in}} \cdot \bm{W}_{\mathrm{out}}^{\mathrm{f}} \bm{u}^{\mathrm{f}}(k)
+ \bm{W}_{\mathrm{param}} \cdot I_{\mathrm{fast}}(n)
+ \bm{b}^{\mathrm{f}}) .
\end{eqnarray}
\end{widetext}
The calculation of the LLE follows the standard approach using continuous Gram-Schmidt orthonormalization of the fundamental solutions to the linearized differential equation along the trajectory (\cite{ShimadaNagashima}), which is given by:
\begin{eqnarray}
\label{linear_ode}
\delta \bm{u}(k+1) = J \cdot \delta\bm{u}(k),
\end{eqnarray}
where $J$ is the Jacobian matrix of equation \ref{noy}, expressed as:
\begin{eqnarray}
J =    \dfrac{\partial \bm{u}^{ \mathrm{f}}(k+1)}{\partial\bm{u}^{\mathrm{f}}(k)}  .
\end{eqnarray}
Let $\bm{r}(k)$ be the argument of the hyperbolic tangent function in the RHS of equation \ref{noy}:
\begin{widetext}
\begin{eqnarray}
\bm{r}(k) = \bm{W}^{\mathrm{f}} \bm{u}^{\mathrm{f}}(k) + \bm{W}^{\mathrm{f}}_{\text{in}} \cdot \bm{W}_{\mathrm{out}}^{\mathrm{f}} \bm{u}^{\mathrm{f}}(k)
+ \bm{W}_{\mathrm{param}} \cdot  I_{\mathrm{fast}}(n)
+ \bm{b}^{\mathrm{f}} .
\end{eqnarray}
\end{widetext}
Then, the Jacobian matrix can be described as follows:
\begin{widetext}
\begin{eqnarray}
      \dfrac{\partial \bm{u}^{ \mathrm{f}}(k+1)}{\partial\bm{u}^{\mathrm{f}}(k)}  &=& \left( \bm{E} -\mathrm{diag}\left[\tanh^2  \left(\bm{r}(k)\right) \right] \right)
      \cdot  
      \left(
      \bm{W}^{\mathrm{f}} + \bm{W}^{\mathrm{f}}_{\text{in}} \cdot \bm{W}_{\mathrm{out}}^{\mathrm{f}}
      \right).
\end{eqnarray}
\end{widetext}

\subsection{Data generation models and observation}
We generated time series to train the reservoir model using a nonlinear differential equation whose solutions were computed by numerical integration and discretized at specific time intervals $\Delta t$, with characteristic values for each model.
\subsubsection{Lorenz equation}
As an example of the function $f_x$ described in the problem setting, we used the Lorenz 63 model to define a non-stationary signal source:
\begin{eqnarray}
\label{LorenzSystem}
\left\{ 
  \begin{alignedat}{2}   
    \frac{\mathrm{d}x_1}{\mathrm{d}t} &= a(x_2 - x_1) \\
    \frac{\mathrm{d}x_2}{\mathrm{d}t} &= -x_2 + x_1 (\lambda - x_3) \\
    \frac{\mathrm{d}x_3}{\mathrm{d}t} &= -bx_3 + x_1 x_2
  \end{alignedat} 
\right.
\end{eqnarray}
where $a$ and $b$ are parameters, and $\lambda$ is considered to change slowly over time.  The observation time series $y(n)$ is given by:

\begin{eqnarray}
    y(n) =   x_1(\Delta t \cdot n ) 
\end{eqnarray}
where $n \in \mathbb{N}$ is the index of the discretized time steps. We set the parameter values to $a=10$ and $b=8/3$, which are commonly employed, and discretized the time series with a time step of $\Delta t = 0.05$.

\subsubsection{R\"ossler equation}
The R\"ossler equation was also utilized as a data generation model:
\begin{eqnarray}
\left\{ 
\begin{aligned}
\frac{\mathrm{d}x_1}{\mathrm{d}t} &= -x_2 - x_3 \\
\frac{\mathrm{d}x_2}{\mathrm{d}t} &= x_1 + ax_2 \\
\frac{\mathrm{d}x_3}{\mathrm{d}t} &= \lambda + x_3(x_1 - c)
\end{aligned} 
\right.
\end{eqnarray}
where $a$ and $c$ are static parameters, and $\lambda$ is considered to change slowly over time. The observation time series is $y(n) = x_1(\Delta t \cdot n ) $ as in the case of the Lorenz system, and the parameter values are set to $a = 0.2, c =5.7,$ and $\Delta t = 0.7$.

\section{Results}

\begin{figure*}
    \centering
    \includegraphics[width=134 mm]{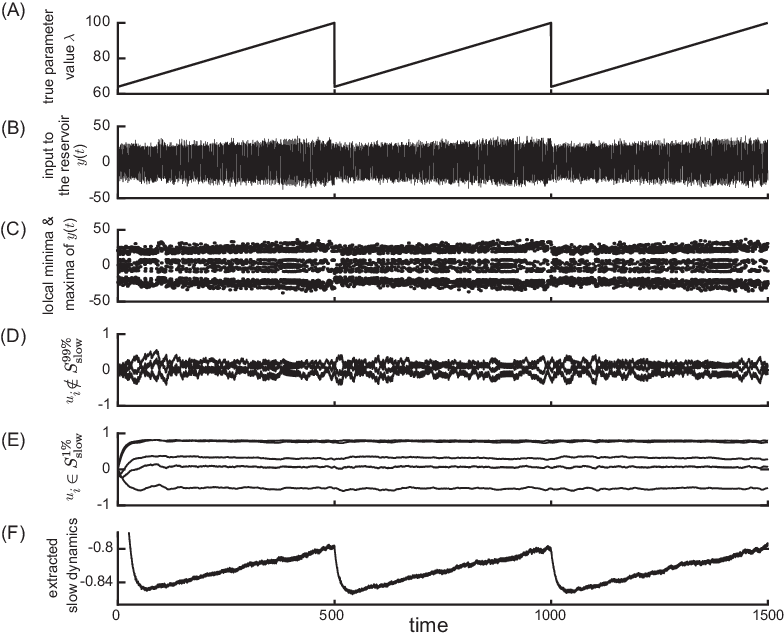}
    \caption{Response of the slow reservoir to time series generated by the Lorenz system in the Experiment 1. (A) True parameter value $\lambda$ of the Lorenz system slowly changing from $\lambda = 64$ to $\lambda=100$.(B) Variable $y(n) = x_1(\Delta t \cdot n)$, representing the first element of the state of the Lorenz system used as the input to the reservoir. (C) Local minima and maxima of the trace shown in (B). (D, E) Values of the internal states, $x_i$, of the slow reservoir characterized by rapid and slow temporal fluctuations, respectively. (F) Extracted slow dynamics calculated as the average of the absolute values of internal nodes exhibiting slow behavior. All panels are plotted against time in the horizontal axis.}
    \label{Result_Lorenz}
\end{figure*}
\subsection{Experiment 1: Extraction of slow features from time series using the slow reservoir}\label{esf}
\begin{figure*}
    \centering
    \includegraphics[width=134 mm]{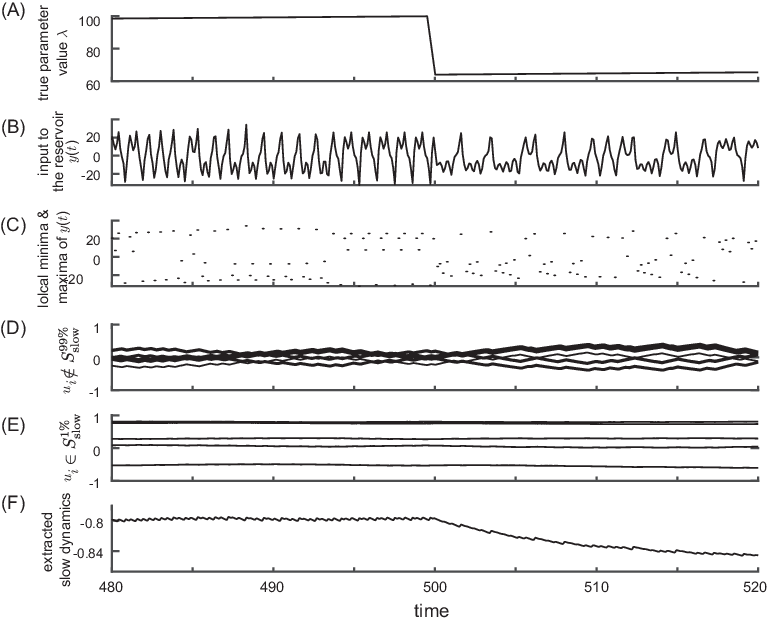}
    \caption{Response of the slow reservoir to time series generated by the Lorenz system in the Experiment 1 with an expanded time axis. This figure shows the same data as in fig. \ref{Result_Lorenz} but with an expanded time scale. (A) True parameter value $\lambda$ of the Lorenz system. (B) Variable $y(n) = x_1(\Delta t \cdot n)$, representing the first element of the state of the Lorenz system used as the input to the reservoir. (C) Local minima and maxima of the trace shown in (B). (D,E) Values of the internal states, $x_i$, of the slow reservoir characterized by rapid and slow temporal fluctuations, respectively. (F) Extracted slow dynamics calculated as the average of the absolute values of internal nodes exhibiting slow behavior. All panels are plotted against time in the horizontal axis.}
    \label{Result_Lorenz_zoom}
\end{figure*}

Initially, we investigated the feasibility of observing parameter dynamics within a reservoir by feeding observed time series data from a nonlinear system with slowly changing parameter values into a reservoir characterized by a slow time constant. A schematic overview of the numerical computations performed in this study is depicted in fig.~\ref{schematic1}. The response of the slow reservoir to time series generated from the Lorenz system is illustrated in fig. \ref{Result_Lorenz}, with a more detailed view provided in fig. \ref{Result_Lorenz_zoom} using a shorter time scale. In the Lorenz system described by equation \ref{LorenzSystem}, the parameter $\lambda$ varies slowly over time, following a triangular wave pattern between \(\lambda = 64\) and \(\lambda = 100\) (fig.~\ref{Result_Lorenz}(A)). Notably, the period of change in the parameter \(\lambda\) is 500, which is two orders of magnitude larger than the typical timescale of the Lorenz attractor ($\approx 1$). The time series of the variable \(x_1\) reflects these variations in parameter values, as shown in fig.~\ref{Result_Lorenz} (B-C). Figure \ref{Result_Lorenz_zoom} presents the same data as fig.~\ref{Result_Lorenz} but with a modified horizontal axis scale. At approximately \(t=500\), an abrupt change in the parameter value leads to a significant alteration in the shape of the \(x_1\) time series, as indicated in fig.~\ref{Result_Lorenz_zoom} (B-C). Upon examining the internal state of the slow reservoir, we observed that certain nodes exhibited rapid temporal fluctuations (fig.~\ref{Result_Lorenz} (D) and fig.~\ref{Result_Lorenz_zoom} (D)), while others displayed slower activities characterized by minimal high-frequency components (fig.~\ref{Result_Lorenz} (E) and fig.~\ref{Result_Lorenz_zoom} (E)). 

Our objective is to extract patterns of parameter fluctuations from the internal state of the slow reservoir in an unsupervised manner, assuming that the parameter's fluctuation is slower than the typical timescale of the time series. To accomplish this, we identified nodes exhibiting slow changes (see Materials and Methods).  The nodes with the ten highest SD values are depicted in fig.~\ref{Result_Lorenz} (D) and fig.~\ref{Result_Lorenz_zoom} (D), while those with the five lowest values are shown in fig.~\ref{Result_Lorenz} (E) and fig.~\ref{Result_Lorenz_zoom} (E). Furthermore, we selected the 10\% of nodes, i.e., 50 nodes, with the smallest SD values and calculated the negative mean of their absolute values, as illustrated in fig.~\ref{Result_Lorenz} (F) and fig.~\ref{Result_Lorenz_zoom} (F) (see Materiels and methods for details). As shown in fig.~\ref{Result_Lorenz}(F), the average activity of the extracted slow nodes follows a trend similar to the temporal variation of the parameter $\lambda$. Given that approximately half of the nodes exhibit an inverted pattern, we took the absolute values of each node's values before averaging to ensure consistent directionality. The negative value of the final values is presented for easier comparison with the parameter $\lambda$’s fluctuation pattern. It's important to note that this process of taking the negative value, which relies on knowledge of the true parameter value, is unnecessary for predicting the unknown bifurcation presented in Subsection \ref{pub}.

\begin{figure*}
    \centering
    \includegraphics[width=134 mm]{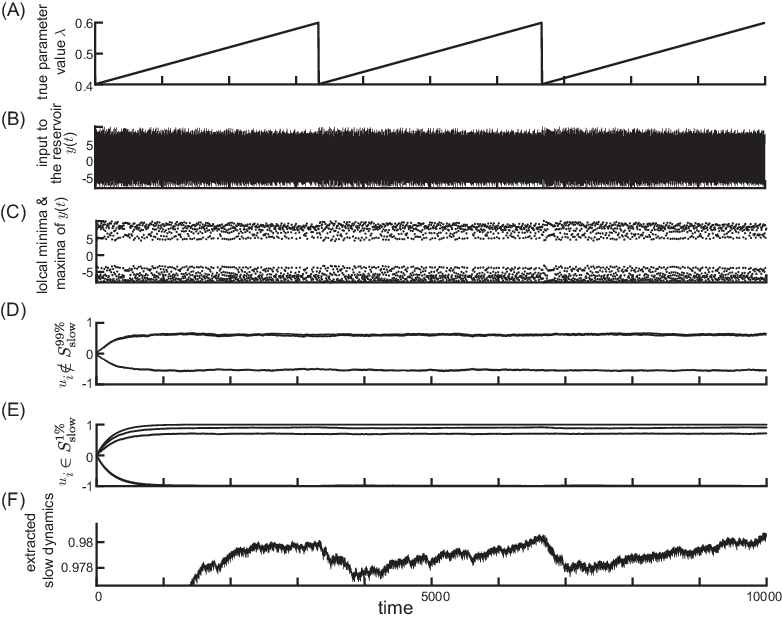}
    \caption{Response of the slow reservoir to time series generated by the R\"ossler equation. (A) True parameter value $\lambda$ of the R\"ossler equation. (B) Variable $y(n) = x_1(\Delta t \cdot n)$, representing the first element of the state of the R\"ossler equation used as the input to the reservoir. (C) Local minima and maxima of the trace shown in (B). (D, E) Values of the internal states, $x_i$, of the slow reservoir characterized by rapid and slow temporal fluctuations, respectively. (F) Extracted slow dynamics calculated as the average of the absolute values of internal nodes exhibiting slow behavior. All panels are plotted against time in the horizontal axis.}
    \label{Result_Roessler}
\end{figure*}

\begin{figure*}
    \centering
    \includegraphics[width=134 mm]{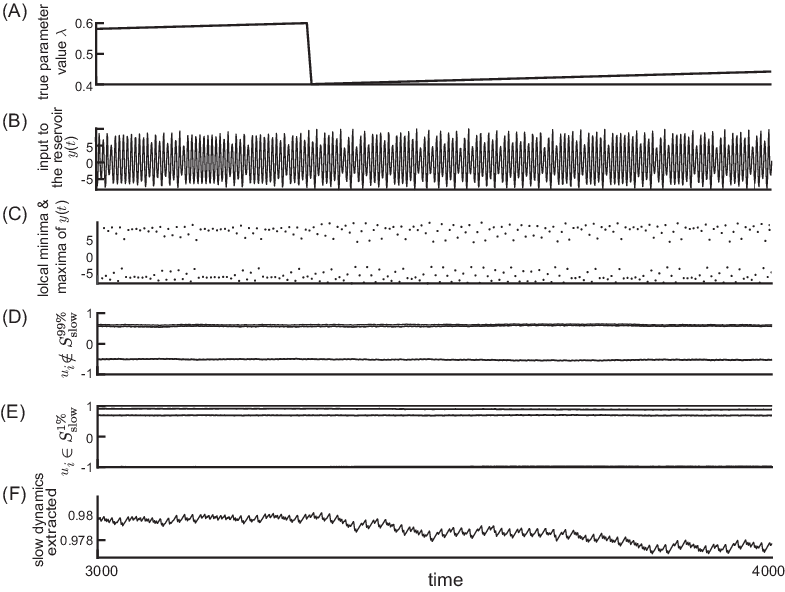}
    \caption{Response of the slow reservoir to time series generated by the R\"ossler equation with an expanded time axis. This figure shows the same data as in fig. \ref{Result_Roessler} but with an expanded time scale. (A) True parameter value $\lambda$ of the R\"ossler equation. (B) Variable $y(n) = x_1(\Delta t \cdot n)$, representing the first element of the state of the R\"ossler equation used as the input to the reservoir. (C) Local minima and maxima of the trace shown in (B). (D,E) Values of the internal states, $x_i$, of the slow reservoir characterized by rapid and slow temporal fluctuations, respectively. (F) Extracted slow dynamics calculated as the average of the absolute values of internal nodes exhibiting slow behavior. All panels are plotted against time in the horizontal axis.}
    \label{Result_Roessler_zoom}
\end{figure*}

Figures \ref{Result_Roessler} and \ref{Result_Roessler_zoom} show results of a similar analysis using the time series generated from the R\"ossler equation as input. Although the extracted slow dynamics in the R\"ossler attractor do not distinctly exhibit parameter variations as in the Lorenz attractor, the shape of the parameter variations remains observable, as demonstrated in fig.~\ref{Result_Roessler} (F) and fig.~\ref{Result_Roessler_zoom} (F).

Within the framework of reservoir computing, regression to the training data from the internal states of the reservoir is a common practice. Based on the obtained results, it is clear that supervised learning regression can be applied to the time series of the parameter $\lambda$ from the internal state of the slow reservoir (the results of supervised fitting are shown in Supplementary Figure 2). However, even without such supervised learning, if one can assume foresightedly that "the parameter variations have a much slower timescale than the typical timescale of the observed time series, allowing for the separation of timescales," then, as demonstrated in this study, it might be possible to estimate the pattern of parameter variations simply by observing the activity of slowly moving nodes within the reservoir's internal state. 

\begin{figure*}
    \centering
    \includegraphics[width=134 mm]{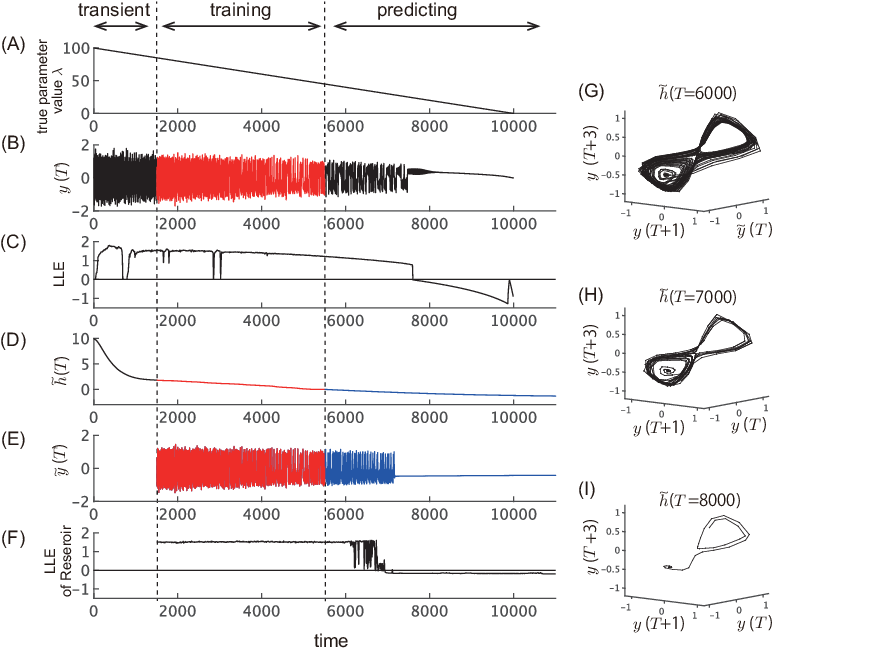}
    \caption{Prediction of unobserved bifurcation. This figure presents the time series during consecutive training and predicting phases. (A-C) Time evolution of the Lorenz system that generates the data to be learned by the model. (A) Slowly varying true parameter value $\lambda$ of the Lorenz system plotted against time. (B) Time series generated by the Lorenz system, $y(n) = x_1(\Delta t \cdot n )$, where $x_1(t)$ is the first variable of the Lorenz system. The time series from $n=1500$ to $n=5500$ is used as training data and fed into the model. Bifurcation occurs around $n=7500$, where the chaotic oscillation vanishes. After this point, the trajectory converges to a stable fixed point with one real eigenvalue and two complex conjugate eigenvalues. (C) LLE of the Lorenz system corresponding to the value of $\lambda$ shown in (A). The LLE is plotted against time along the horizontal axis, with each value calculated with a static value of $\lambda$ corresponding to the same time point in (A). (D-F) Model outputs. From $n=0$ to $n=5500$, the model is driven by the external input $y(n)$ shown in (B). After $n=5500$, the model switches to the closed-loop model depicted in fig.~\ref{schematic_unobserved}. (D) Slow dynamics extracted by the slow reservoir ($n=0$ to $n=5500$) and prediction of its dynamics by the slow dynamics predictor after $n=5500$ (blue line). The extracted slow dynamics before $n=1500$ are treated as transient dynamics and not used as training data. (E) Output of the fast reservoir. The red line plotted between $n=1500$ and $n=5500$ shows the fitting of the training data shown in (B). The blue line plotted after $n=5500$ shows the prediction of the data shown in (B) by the closed-loop model, evidencing the bifurcation from chaotic oscillation to oscillation death observed in the original Lorenz model shown in (B). (F) LLE of the closed-loop fast reservoir, calculated with a fixed value of the external input $p$. The value of $p$ is set to the value of $h(n)$ or $\tilde{h}(n)$ plotted at the same time point in (D). (G-I) Reconstructed attractor shape by delay embedding of the fast reservoir output $y(n)$ with the input to the fast reservoir at $\tilde{h}(n=6000),\ \tilde{h}(7000)$, and $\tilde{h}(8000)$. The trajectories converge to a stable fixed point in (H) and (I).}
    \label{unobserved}
\end{figure*}

\subsection{Experiment 2: Prediction of unobserved bifurcation from time series data}\label{pub}

As demonstrated in the previous subsection, extracting time series of slowly-varying elements from the slow reservoir allows us to unsupervisedly reveal the underlying slow parameter dynamics of the target system. As discussed in the Introduction, Patel et al. (\cite{patel2021using}) and Kim et al. (\cite{kim2021teaching}) have shown that by concurrently inputting the time series of actual parameter values into the reservoir, it is possible to predict bifurcations in the system’s attractors, even if these bifurcations are not present in the training data. In this case, we present a scenario where the slow dynamics, unsupervisedly extracted by the slow reservoir, are fed into the reservoir separately from the observations of the fast dynamics, enabling the prediction of bifurcations in the target system that are not contained in the training data. Our model, depicted in Figure \ref{schematic_unobserved}, consists of three reservoirs: the slow reservoir, the slow dynamics predictor, and the fast reservoir. The slow reservoir is characterized by long time constants in its leaky units and a spectral radius equal to 1, specifically engineered to extract the slowest-moving dynamics by calculating the mean absolute values of the 10\% most slowly changing elements. The slow dynamics predictor receives outputs from the slow reservoir and learns its dynamics. The output from this slow reservoir is smoothed through a linear filter before being sent to the two downstream reservoirs (Methods). Meanwhile, the fast reservoir receives both the fast dynamics directly observed from the target system and the slow dynamics extracted by the slow reservoir, predicting the subsequent state of $y(n)$. We use time series data generated from the Lorenz attractor with a slowly varying parameter as the target system to learn. Figure \ref{unobserved} illustrates the results of learning by the model. 

Figure \ref{unobserved} illustrates the results of the model's learning process. In this numerical experiment, the parameter $\lambda$ changes linearly and gradually over time (fig.~\ref{unobserved}) under a scenario where bifurcations occur in the fast dynamics leading to the disappearance of the chaotic attractor (fig.~\ref{unobserved}). Beyond $n=7500$, the Lorenz system develops a stable fixed point characterized by two complex conjugate eigenvalues. As shown in fig.~\ref{unobserved}(B), chaotic oscillations occur prior to $n=7500$, but suddenly cease at a specific point. After $n=7500$, the system converges towards one of the two stable fixed points. Figure \ref{unobserved}(C) depicts the value of the LLE for the Lorenz system when the parameter value displayed in fig.~\ref{unobserved}(A) remains constant over time. Note that this LLE is computed for each point along the horizontal axis with a fixed value of $\lambda$, unlike in a system with a temporally varying $\lambda$. As shown in fig.~\ref{schematic_unobserved}(A), the model operates as an open-loop model from $n=0$ to $n=5500$, driven by the external input $y(n)$ depicted in fig.~\ref{unobserved}(B). The slow dynamics extracted from the reservoir after smoothing by the linear filter are presented in fig. \ref{unobserved}(D). The model undergoes a transient phase up to $n=1000$, but then exhibits nearly linear behavior resembling the true parameter variations shown in (A). Starting at $n=5500$, the slow dynamics predictor is changed to a closed-loop model, generating $\tilde{h}(n)$, the prediction of $h(n)$. At this point, the internal state of the reservoir, determined by the external force during the training phase, remains unchanged; only the input is instantaneously replaced by the feedback from its own output. Figure \ref{unobserved}(E) shows the output from the fast reservoir. The results of fitting $y(n)$ (the time series shown in (B)) from $n=1500$ to $n=5500$ are marked in red. After $n=5500$, $\tilde{h}(n)$ generated by the slow dynamics predictor, along with the feedback from its own output, are fed to the fast reservoir to generate predictions for $y(n)$. At this stage, the entire system operates as an autonomous system with no external input. By $n=7000$, chaotic oscillations disappear, and the trajectory converges to a stable fixed point, as shown in fig.~\ref{unobserved}(E).

For each time point along the horizontal axis in fig.~\ref{unobserved}(E), the LLE of the closed-loop fast reservoir is estimated by fixing the value of the input $h(n)$ to that at each time point in panel (D). It was found that, during the chaotic oscillations observed in the fast reservoir's output before $n=6000$, the LLE closely matches that of the Lorenz system during the learning phase (fig.~\ref{unobserved}(F)). However, the system still does not fully replicate the slight decreasing tendency of the LLE and the narrow windows with zero LLE found in the original Lorenz system. After $n=7000$, the LLE of the fast reservoir takes negative values. Using the fixed values of the input to the fast reservoir, $\tilde{h}(n=6000)$, the attractor reconstructed by the delay time coordinate shows a shape akin to that of the Lorenz attractor (fig.~\ref{unobserved}(G)).  Furthermore, when using the values of $\tilde{h}(n)$ after the oscillations have ceased, the trajectory converges with rotation around a single stable fixed point (fig.~\ref{unobserved}(H, I)). This suggests that the existence in the original Lorenz system of a stable fixed point with one real eigenvalue and two conjugate complex eigenvalues is predicted only by learning from the chaotic time series as $\lambda$ decreases. Overall, the results shown in fig.~\ref{unobserved} suggest that, solely by observing the fast variable, the model successfully learned the dynamical flow of the original Lorenz system, including the shape of the attractor, its stability, and its slowly changing vector field.

\section{Discussion}

We have demonstrated that unsupervised extraction of the very slowly changing parameters of the dynamical system generating the signals is possible by simply feeding the observation to a reservoir with a long-time scale and selecting the internal nodes of the reservoir with slowly varying states. Furthermore, we have shown this reservoir's capability to predict bifurcations not present in the training data, such as the death of chaotic oscillations, by inputting the extracted slow features and observation signal into another reservoir. Kim et al. (\cite{kim2021teaching}) and Patel et al. (\cite{patel2021using}) demonstrated the prediction of unobserved bifurcations not present in the training data using a reservoir computing framework. Their work illustrated the remarkable capability of reservoir computing to learn the parameter dependencies within dynamical system flows and to reproduce unknown bifurcations. However, they treated the parameters as known, which is not the case in real-world applications, where the values of these parameters often cannot be observed. In this study, we introduce two reservoirs: a slow feature predictor that forecasts the movement of these slow features, and a fast reservoir that predicts the values of the observed time series. By inputting the slow features resulting from the unsupervised extraction, we establish a closed-loop model that operates as a fully autonomous dynamical system during the predicting phase. This demonstrates the ability to forecast the emergence of unknown bifurcations without any direct observation of the parameter value. Nonlinear, non-stationary processes are abundant in various natural and physical phenomena. Additionally, numerous scenarios probably exist where slow dynamics inducing qualitative dynamics changes remain unobservable. The potential applications of this approach are vast, spanning fields such as neuroscience (including electrophysiological measurements, electroencephalography, functional Magnetic Resonance Imaging, and disease progression with tipping points), material science (including surface science), and weather prediction and control. 

The main limitation of the current work is the lack of understanding of the principle underlying the phenomenon wherein the behavior of slowly moving nodes, selected heuristically from within the slow reservoir, is similar to variations in the original system's parameters. As mentioned in the Introduction, the observations made could be explained if the reservoir can achieve generalized synchronization with the target system (\cite{Caroll_10.1063/1.5128898, PhysRevE.51.980}), including the slow parameter dynamics. For the results shown in fig.~\ref{schematic1}, we conducted the same numerical simulation using a reservoir with linear dynamics by replacing the activation function in eq.~\ref{slowreservoirequ} with the identity map. The results show that a linear reservoir with a slow time constant does not yield parameter estimation, even with supervised training, where the internal state of the reservoir is fitted to the true parameter value (fig.~S2). This suggests that the nonlinearity of the reservoir is crucial for the current results. However, generalized synchronization would not fully explain the current results. For example, fast-moving nodes also exist within the slow reservoir. It is not trivial that in the internal state of the reservoir, $\bm{u}^{\mathrm{s}} \in \mathbb{R}^{N^{\mathrm{s}}}$, the directions of fast and slow fluctuations align along axes, $u_1, u_2, \cdots, u_{{N^{\mathrm{s}}}}$ (i.e., different nodes). In fact, it would not be surprising if fast and slow fluctuations were superimposed at all single nodes. Therefore, further investigation is required to elucidate the logical reason why simply selecting slow-moving nodes worked well as a heuristic. Conversely, employing a more sophisticated method to separate the directions of fast and slow fluctuations might lead to better performance (\cite{ANTONELO2012178}).

Previous works have extensively explored the behavior of complex systems around tipping points (\cite{Dakos2008,Veraart2011, Liu2013}). For instance, the Dynamical Network Biomarker (DNB) method captures the increase in temporal fluctuations and the intensified correlation associated with critical slowing down (\cite{Liu2013}). Unlike the approach in the current study, which involves learning the flow of the dynamical system in a relatively low-dimensional, deterministic, and strongly nonlinear phase space, the DNB method utilizes the generic behavior near bifurcation points in very high-dimensional systems based on linearization around a fixed point. Given their distinct advantages, combining these methods in the future might improve the prediction and control of non-stationary, nonlinear systems. 

Kim et al. the emergence of chaotic attractors in the Lorenz system by extrapolating the parameter space and learning in regions without chaotic attractors, where only two stable fixed points exist (\cite{kim2021teaching}). In our research, we have extracted the slowly changing parameters of the target system by receiving its generated time series through the reservoir. However, applying our method to predict the emergence of a chaotic strange attractor  by learning observation from the Lorenz system with stable fixed points is currently challenging because our method relies on observing long time series to extract slow features, whereas the target system does not produce a long time series with oscillations if the trajectory converges to a fixed point. A new framework would be necessary to estimate the parameter changes in a system with stable fixed points, e.g., by introducing external perturbations to the target system and receiving its response through the reservoir.

\section{Author Contributions}
K.T. and Y.K. conceived the design of the study. K.T. performed numerical simulations, and analyzed the data. K.T. and Y.K. wrote the paper. All authors contributed to the discussion and approved the paper.

\section{Acknowledgments}
This work was supported by JSPS KAKENHI (Nos. 20K19882, 20H04258, 20H00596, 21H05163, 23K11259, 23H03468, 24H02330) and the Japan Science and Technology Agency (JST) Moonshot R\&D (JPMJMS2284, JPMJMS2389), JST CREST (JPMJCR17A4), JST ALCA-Next  (JPMJAN23F3). This paper is also based on results obtained from a project JPNP16007 commissioned by the New Energy and Industrial Technology Development Organization (NEDO).

\bibliography{ref}

\newpage

\section{Supplementary Tables and Figures}





Figure~S1 shows the slow reservoir's output, $\tilde{u}^s(n)$ and the smoothed slow reservoir's output filtered by linear dynamics, $h(n)$. The filtering procedure did not affect the waveform of $h(n)$.

\begin{figure*}[htbp]
\begin{center}
\includegraphics[width=14cm]{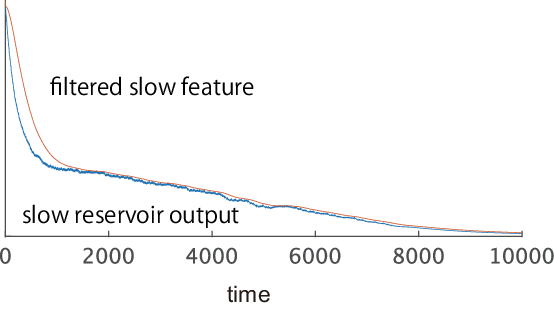}
\end{center}
\caption{The slow reservoir's output and smoothed input to the downstream reservoirs. The application of a linear filter does not significantly alter the shape of the time series; it is used solely for removing high-frequency components and smoothing.}
\label{figs:1}
\end{figure*}

\begin{figure*}[htbp]
\begin{center}
\includegraphics[width=14cm]{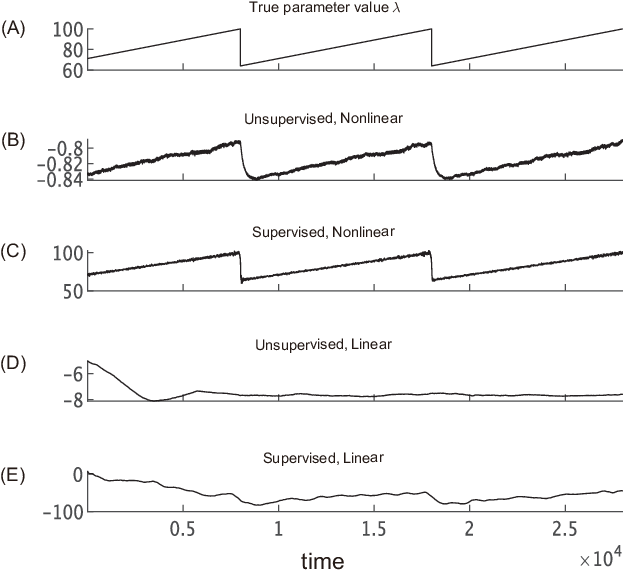}
\end{center}
\caption{The slow reservoir's output and smoothed input to the downstream reservoirs. The application of a linear filter does not significantly alter the shape of the time series; it is used solely for removing high-frequency components and smoothing.}
\label{figs:2}
\end{figure*}

To investigate whether nonlinearity, as well as the slow timescales of the reservoir, is crucial for the extraction of slow features, we conducted identical computations using a linear reservoir in which the activation function $\tanh$ in equation 8 in the main text is replaced with the identity map (fig.~\ref{figs:2}). Figure~\ref{figs:2}(B) shows the results of slow feature extraction by the nonlinear slow reservoir, which follows a pattern similar to the true parameter variations depicted in fig.~\ref{figs:2}(A), which is the same result as in the main text. In contrast, when the reservoir's activation function $\tanh$ is replaced with an identity map, the outcomes, as shown in fig.~\ref{figs:2}(D), do not correlate with the true parameter variations shown in fig.~\ref{figs:2}(A). Furthermore, to more directly verify whether the internal states of the reservoir has information about the slowly varying true parameter values, we fitted the internal states of the reservoir to the true parameter values $\lambda$ shown in fig.~\ref{figs:2}(A). Namely, the common reservoir computing framework setting for readout fitting is done. The results, depicted in fig.~\ref{figs:2}(C), reveal that fitting yields small residuals between the result and the true parameter value. However, with the linear model, even when fitting is performed in a supervised manner, fails to replicate the changes in the parameter $\lambda$ (fig.~\ref{figs:2}(E)). Therefore, the results obtained suggest that the nonlinearity of the slow reservoir is necessary.

\end{document}